\documentclass[11pt]{article}
\usepackage{amsmath} 
\usepackage[final]{acl}

\usepackage{times}
\usepackage{latexsym}
\usepackage{booktabs}
\usepackage{algorithm}
\usepackage{algpseudocode}
\usepackage{amsmath}
\usepackage[most]{tcolorbox}
\newtcolorbox{promptbox}{
    colback=gray!3,
    colframe=gray!40,
    boxrule=0.35pt,
    arc=1.8mm,
    left=1.5mm,
    right=1.5mm,
    top=0.8mm,
    bottom=0.8mm
}
\usepackage[table]{xcolor}
\usepackage{array}
\usepackage{xcolor}
\definecolor{DeepRed}{HTML}{C00000}
\definecolor{DeepBlue}{HTML}{00509d}  
\usepackage{enumitem}
\setlist[itemize]{leftmargin=*,noitemsep}
\usepackage{multirow}
\usepackage[T1]{fontenc}
\usepackage[utf8]{inputenc}

\usepackage{microtype}

\usepackage{inconsolata}

\usepackage{graphicx}

\title{SABER: Stability-Aware Early Exit for LLM Reasoning via Adversarial Branch Probing
}
\author{
 \textbf{Wanli Cheng\thanks{Equal contribution.}},
 \textbf{Haiya Xiang\footnotemark[1]},
 \textbf{Juntao Li\thanks{Corresponding author.}},
 \textbf{Hongling Wang},
 \textbf{Wenliang Chen}
\\
 Soochow University, China
\\
 \texttt{\{wlchengsmk666,hyxiang\}@stu.suda.edu.cn},
 \texttt{ljt@suda.edu.cn}
}
\begin{document}
\maketitle
\begin{abstract}
Large Reasoning Models (LRMs) achieve strong reasoning capabilities, yet long-chain reasoning becomes inefficient once the intermediate answer stabilizes across reasoning steps: additional reasoning yields little marginal benefit while incurring substantial inference cost. Existing early-exit methods based on confidence or entropy poorly capture reasoning stability, while consistency-based approaches rely on multi-step trajectory agreement, requiring sequential evaluations that delay exit. To better balance efficiency and reliability, we propose SABER, a training-free framework for stability-aware early exit via adversarial branch probing. SABER constructs simple yet effective semantic perturbations around intermediate reasoning states to form adversarial branches, and applies lightweight probing to estimate their likely final outcomes without full trajectory rollouts. When the probed outcomes remain consistent across branches, SABER exits early; otherwise, it continues reasoning. Experiments across multiple reasoning benchmarks and model architectures show that SABER reduces reasoning token consumption by 30.2\%--39.8\% on average while maintaining competitive accuracy with full-length reasoning.\footnote{Code is available at \url{https://github.com/Bl1nding/SABER}.}
% \footnote{To facilitate reproducibility, our code is available at
% \url{https://github.com/Bl1nding/SABER}.}
\end{abstract}

\section{Introduction}
\label{sec:Introduction}

In recent years, Large Reasoning Models (LRMs), such as OpenAI o1~\citep{jaech2024openai}, DeepSeek R1~\citep{guo2025deepseek}, and Qwen3~\citep{yang2025qwen3}, have made remarkable progress on complex tasks, including mathematical reasoning and code generation, by leveraging test-time scaling and long Chain-of-Thought (CoT) reasoning~\citep{snell2024scaling}. However, prior work shows that LRMs often exhibit an overthinking phenomenon, generating redundant verification steps or exploring unnecessary reasoning paths even after sufficient information has been obtained~\citep{chen2024not,sui2025stop}. Such behavior reduces inference efficiency and may introduce superfluous reasoning steps or error accumulation during decoding. These challenges motivate early exit as a promising approach to efficient LRM inference.

To mitigate inference inefficiency caused by long reasoning trajectories, recent work has explored various early-exit strategies for adaptive reasoning. Existing methods such as DEER~\citep{yang2025dynamic} and EAT~\citep{wang2025entropy} determine stopping points using confidence or entropy signals from intermediate reasoning states. However, LRMs often suffer from severe confidence miscalibration, remaining highly confident even when their reasoning is incorrect~\citep{mei2026reasoning,slobodkin2023curious,simhi2025trust,kadavath2022language}.
Beyond confidence-based approaches, another line of work exploits consistency as a stopping signal. Representative methods include Dynasor~\citep{fu2025reasoning}, which leverages agreement among intermediate answers for adaptive stopping.
While more reliable, such methods typically require consistency checks across multiple subsequent reasoning steps, delaying early exit and introducing extra inference overhead. 
Therefore, constructing a stopping signal that can reliably reflect convergence without requiring expensive long-horizon consistency accumulation remains an open challenge.

Answer convergence~\citep{liu2025answer} provides an effective signal for early exit in long-chain reasoning. Building on this insight, we hypothesize that reasoning trajectories approaching convergence exhibit stability under local semantic perturbations, causing perturbed reasoning branches to yield consistent predictions. We therefore propose \textbf{Stability-Aware Early Exit for LLM Reasoning via Adversarial Branch Probing (SABER)}, a lightweight and training-free framework that enables adaptive early exit by estimating reasoning stability across perturbed branches.

Inspired by self-consistency~\cite{wang2022self}, SABER constructs parallel neutral and adversarially perturbed branches from the same intermediate reasoning state, and performs lightweight probing using short answer-focused continuations rather than full trajectory rollouts. Instead of relying solely on single-pass confidence signals or long-horizon consistency accumulation, SABER directly probes local reasoning stability under perturbation. Specifically, SABER estimates reasoning stability from two complementary signals: \textbf{Semantic Consistency (SC)}, which quantifies prediction consistency between neutral and adversarial branches, and \textbf{Confidence Stability (CS)}, which captures the stability of model confidence across branches. These signals are combined into a unified \textbf{Reasoning Stability Score (RSS)} for adaptive early-exit decisions. By jointly modeling semantic and confidence stability, SABER achieves a better trade-off between inference efficiency and reasoning quality.

Our contributions are summarized as follows:
\begin{itemize}

\item We introduce a stability-aware perspective for adaptive early exit, and empirically show that reasoning trajectories leading to correct solutions exhibit substantially stronger stability under semantic perturbations.

\item We propose SABER, a lightweight and training-free early-exit framework that probes reasoning stability through adversarial branching and jointly models semantic consistency and confidence stability for adaptive stopping.

\item Extensive experiments across diverse reasoning benchmarks and model architectures demonstrate that SABER consistently reduces reasoning token consumption by \textbf{30.2\%--39.8\%} while maintaining competitive overall accuracy.

\end{itemize}

\begin{figure}[t]
    \centering
    \includegraphics[width=\linewidth]{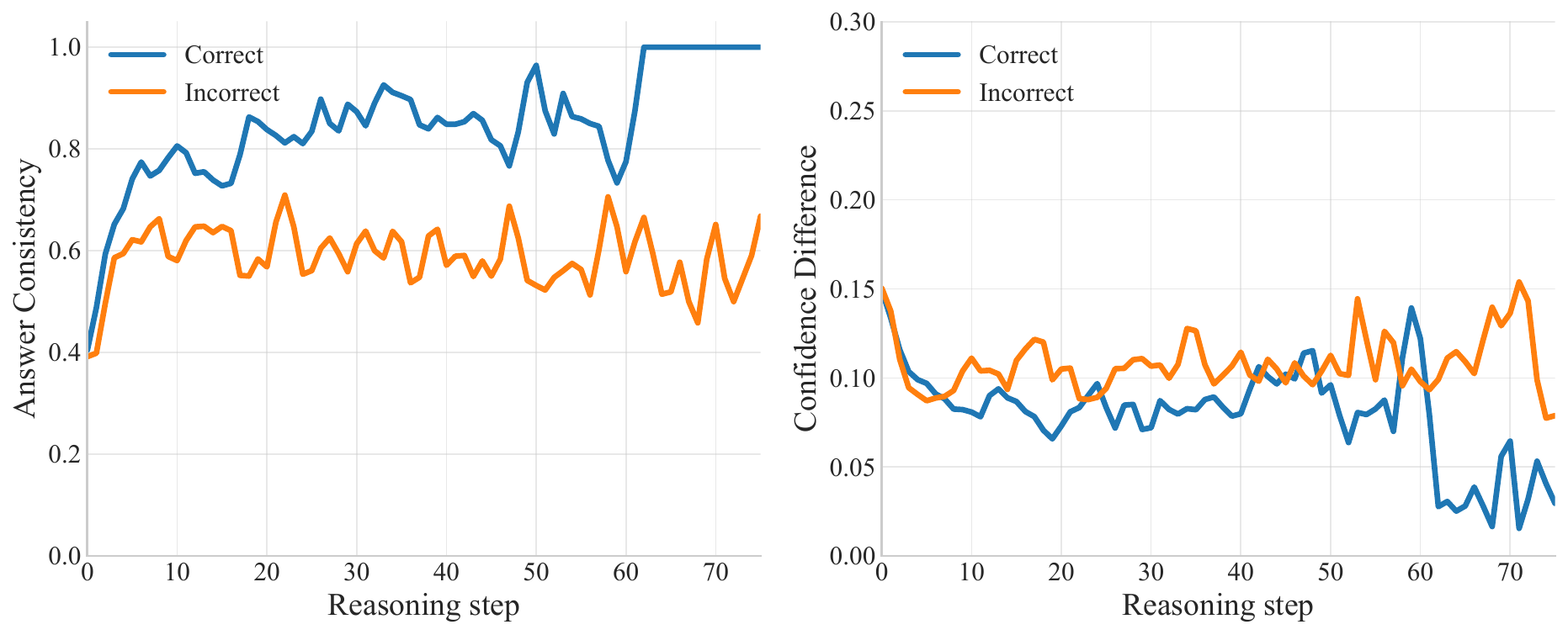}
    \caption{
    Dynamics of reasoning behavior under adversarial perturbations across reasoning steps.
    \textbf{Left:} Answer consistency.
    Trajectories leading to correct solutions progressively become more stable and consistent under perturbation, whereas trajectories leading to incorrect solutions remain unstable.
    \textbf{Right:} Confidence variation.
    Trajectories leading to correct solutions exhibit progressively reduced confidence variation under perturbation, while incorrect trajectories remain highly volatile.
    }
    \label{fig:robustness_components}
\end{figure}

\section{Reasoning Stability under Adversarial Perturbations}
\label{sec:motivation}
\begin{figure*}[t]
    \centering
    \includegraphics[width=\textwidth,trim=10 10 10 10,clip]{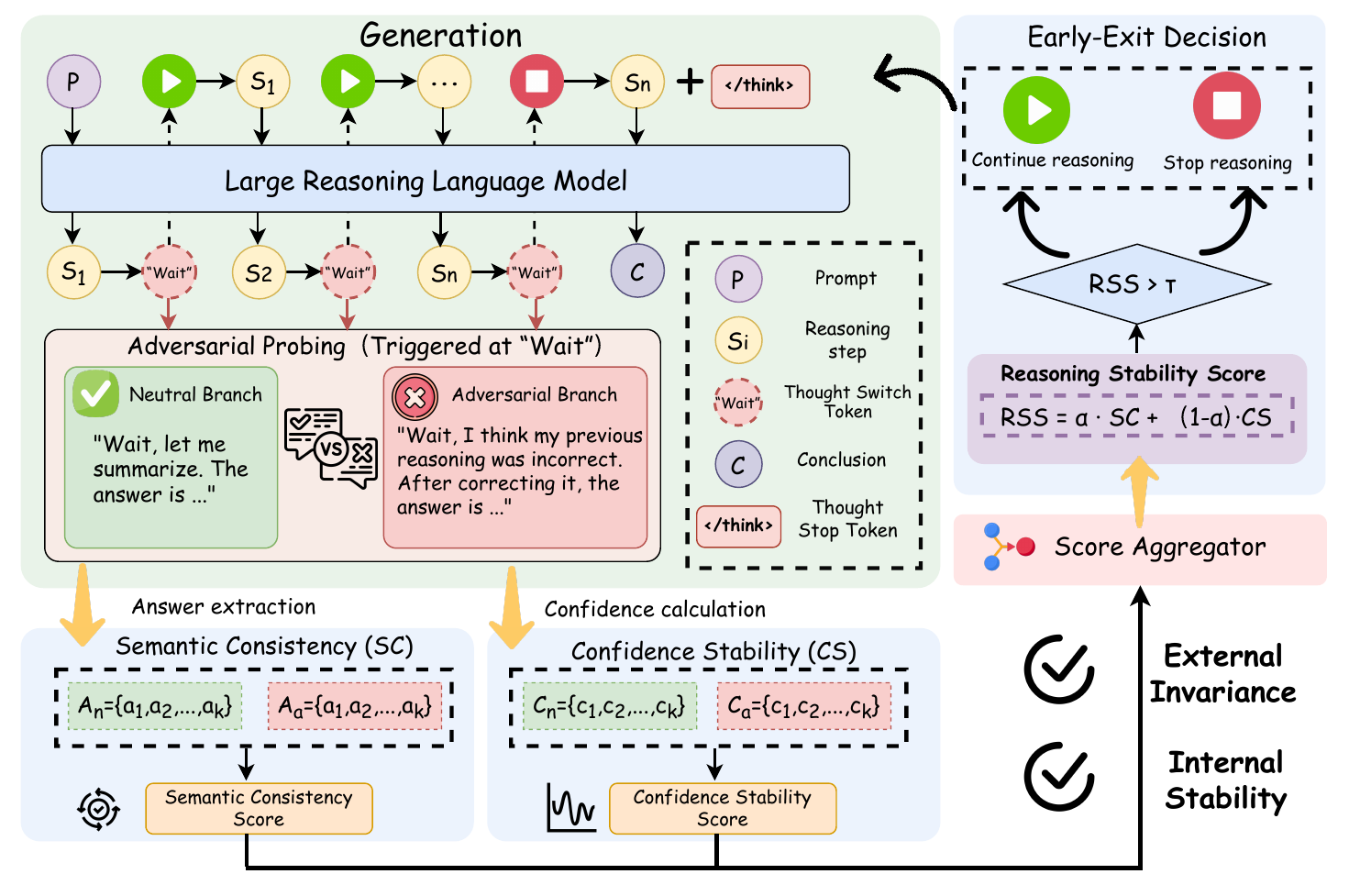}
    \caption{
    Overview of SABER. The framework performs adversarial probing by constructing neutral and adversarial branches at intermediate reasoning steps to estimate reasoning stability under perturbations. Semantic consistency and confidence stability are jointly integrated into the RSS, which dynamically determines whether reasoning should continue or terminate early.
    }
    \label{fig:method}
\end{figure*}

Recent studies suggest that behavioral stability under perturbations provides informative signals for model reliability and uncertainty estimation~\cite{khanmohammadi2025ccps,saadat2026certainty}. Building on these findings, we analyze how intermediate reasoning states evolve under adversarial perturbations throughout multi-step reasoning.

We formalize a reasoning trajectory generated by an LLM as a sequence of tokens
$t_1,t_2,\dots,t_n$.
The trajectory is segmented into intermediate reasoning steps
$S_1,S_2,\dots,S_K$
using transition trigger phrases such as ``Wait'', ``Alternatively'', and
\texttt{``\textbackslash n\textbackslash n''}.
Additional trigger phrases are provided in Appendix~\ref{appendix:rtp}.

At each reasoning step $S_i$, we construct two parallel reasoning branches by appending the following probing prompts to the current reasoning prefix:

\begin{itemize}
    \item \textbf{Neutral Probing Prompt:}
    \emph{``Wait, let me summarize. The answer is \texttt{\textbackslash boxed\{}''}
    
    \item \textbf{Adversarial Probing Prompt:}
    \emph{``Wait, I think my previous reasoning was incorrect. After correcting it, the answer is \texttt{\textbackslash boxed\{}''}
\end{itemize}

The adversarial probe injects misleading corrective signals into the reasoning process, allowing us to examine how intermediate reasoning states respond to perturbations.

For each reasoning step $S_i$, we perform one stochastic sample under both the neutral and adversarial branches to obtain the corresponding predicted answer and confidence score. We conduct adversarial probing throughout reasoning trajectories sampled from OlympiadBench~\cite{olympiadbench} to analyze the evolution of reasoning behaviors under perturbations.

As shown in Figure~\ref{fig:robustness_components}, correct and incorrect reasoning trajectories exhibit markedly different dynamics throughout reasoning. As reasoning progresses, correct trajectories exhibit progressively increasing stability: predictions generated under the two probes increasingly converge, while confidence variations steadily decrease. In contrast, incorrect trajectories remain unstable throughout the reasoning process, exhibiting persistent answer inconsistency and large confidence fluctuations even at later reasoning stages.

These observations indicate that successful reasoning trajectories progressively converge toward stable intermediate reasoning states. This stability consistently manifests in both generated answers and confidence dynamics, providing a potential criterion for characterizing reasoning convergence.

\section{Method}
\label{sec:method}

Motivated by the observation that reasoning trajectories gradually become more stable as the reasoning process approaches a correct solution, we propose SABER, a training-free framework for adaptive early exit based on reasoning stability. As illustrated in Figure~\ref{fig:method}, SABER performs adversarial branch probing at intermediate reasoning steps by constructing neutral and adversarial branches from the same reasoning prefix. Based on the responses from these branches, SABER estimates reasoning stability and aggregates multiple signals into a unified Reasoning Stability Score (RSS), which is used to dynamically determine whether reasoning should continue or terminate early. The detailed formulations of each stability component are provided in the following subsections.

\subsection{Adversarial Branch Probing}

Following the probing setup introduced in Section~\ref{sec:motivation}, SABER performs adversarial probing during reasoning. Specifically, when the token ``Wait'' appears in the reasoning trajectory, the current reasoning prefix is branched into neutral and adversarial continuations.

For each branch, we perform $k$ stochastic decoding samples to obtain answer multisets:
\begin{equation}
\mathcal{A}_{n}=\{a_{n}^{(j)}\}_{j=1}^{k},
\quad
\mathcal{A}_{a}=\{a_{a}^{(j)}\}_{j=1}^{k},
\label{eq:ans}
\end{equation}
where $\mathcal{A}_{n}$ and $\mathcal{A}_{a}$ denote the sampled answers under the neutral and adversarial branches, respectively.

For each sampled continuation, we additionally estimate its generation confidence based on token-level predictive probabilities, yielding two corresponding confidence sets:
\begin{equation}
\mathcal{C}_{n}=\{c_{n}^{(j)}\}_{j=1}^{k},
\quad
\mathcal{C}_{a}=\{c_{a}^{(j)}\}_{j=1}^{k}.
\label{eq:confidence}
\end{equation}
This design enables controlled semantic perturbations of intermediate reasoning states for stability estimation.

\subsection{Reasoning Stability Estimation}

We characterize reasoning trajectory stability from two complementary perspectives: semantic consistency (SC) and confidence stability (CS). Semantic consistency measures the consistency of generated predictions under perturbations, while confidence stability captures the variation of model confidence across perturbed branches. Together, these two signals provide a more reliable estimate of whether the current reasoning state has converged toward a stable solution.

\paragraph{Semantic Consistency.}

We quantify semantic consistency using the multiset Jaccard similarity between predictions from the neutral and adversarial branches:
\begin{equation}
\label{eq:sc}
SC =
\frac{
\sum_{a \in \mathcal{A}_{n} \cup \mathcal{A}_{a}}
\min(f_n(a), f_a(a))
}{
\sum_{a \in \mathcal{A}_{n} \cup \mathcal{A}_{a}}
\max(f_n(a), f_a(a))
},
\end{equation}
where $f_n(a)$ and $f_a(a)$ denote the occurrence frequencies of answer $a$ in $\mathcal{A}_{n}$ and $\mathcal{A}_{a}$, respectively.

\paragraph{Confidence Stability.}

For each sampled continuation, we compute a confidence score as the length-normalized geometric mean of token-level maximum predictive probabilities:
\begin{equation}
c_j =
\left(
\prod_{t=1}^{T}
\max_{v \in \mathcal{V}}
P(v \mid w_{<t})
\right)^{1/T},
\end{equation}
where $\mathcal{V}$ denotes the vocabulary, $w_{<t}$ represents the preceding token sequence, and $T$ is the length of the generated continuation.

The average confidence under the neutral and adversarial branches is then computed as:
\begin{equation}
\bar{P}_n =
\frac{1}{k}
\sum_{j=1}^{k} c_n^{(j)},
\quad
\bar{P}_a =
\frac{1}{k}
\sum_{j=1}^{k} c_a^{(j)}.
\end{equation}

The confidence stability score is defined as:
\begin{equation}
CS = e^{-\gamma \cdot |\bar{P}_n - \bar{P}_a|},
\label{eq:cs}
\end{equation}
where $\gamma$ controls sensitivity to confidence fluctuations. The exponential form amplifies instability under large deviations.

Higher SC and CS jointly indicate greater stability of the reasoning trajectory under adversarial perturbations, suggesting that further reasoning is unlikely to substantially change the final prediction.

\subsection{Early-Exit Decision}

We combine SC and CS into a unified score that reflects reasoning stability:
\begin{equation}
RSS = \alpha \cdot SC + (1 - \alpha) \cdot CS,
\label{eq:rss}
\end{equation}
where $\alpha \in [0,1]$ controls the relative contribution of the two signals. The resulting RSS serves as a unified indicator of reasoning stability and convergence under adversarial perturbations.

During reasoning, SABER continuously monitors the RSS at intermediate reasoning steps. Once the score exceeds a predefined threshold $\tau$, the framework terminates the reasoning process and directly triggers final answer generation. For reasoning models with explicit reasoning delimiters, reasoning is terminated by injecting the closure token \texttt{</think>}.

The underlying intuition is that reasoning trajectories approaching correct solutions tend to become increasingly stable under perturbation. Once the reasoning state reaches a sufficiently stable regime, additional reasoning steps are unlikely to substantially change the final prediction. SABER therefore enables adaptive early exit while largely preserving reasoning quality.

\section{Experiments}
\label{sec:experiments}
\subsection{Experimental Setup}
\label{subsec:setup}

\begin{table*}[t]
\centering
\small
\setlength{\tabcolsep}{3.7pt} % 调整列间距
\renewcommand{\arraystretch}{1.2}

\begin{tabular}{l cc cc cc cc cc cc| cc}
\toprule
 & \multicolumn{10}{c}{\textbf{MATH}} & \multicolumn{2}{c}{\textbf{SCIENCE}}  \\
\multicolumn{1}{c|}{\textbf{Method}} & \multicolumn{2}{c}{\textbf{GSM8K}} & \multicolumn{2}{c}{\textbf{MATH-500}} & \multicolumn{2}{c}{\textbf{AMC23}} & \multicolumn{2}{c}{\textbf{Olympiad}} & \multicolumn{2}{c|}{\textbf{AIME24,25}} & \multicolumn{2}{c|}{\textbf{GPQA-D}} & \multicolumn{2}{c}{\textbf{Overall}} \\
 & Acc$\uparrow$ & Tok$\downarrow$ & Acc$\uparrow$ & Tok$\downarrow$ & Acc$\uparrow$ & Tok$\downarrow$ & Acc$\uparrow$ & Tok$\downarrow$ & Acc$\uparrow$ & Tok$\downarrow$ & Acc$\uparrow$ & \multicolumn{1}{c}{Tok$\downarrow$} &  Acc$\uparrow$ & CR$\downarrow$  \\

\specialrule{0.05pt}{1pt}{1pt}

% 模型区块 1 - 居中并带有背景色
\rowcolor{gray!8}
\multicolumn{15}{l}{\textit{\textbf{DeepSeek-R1-Distill-Qwen-7B}}} \\
Vanilla & \textbf{91.2} & 1,164 & 90.4 & 3,797 & \textbf{90.6} & 6,032 & \textbf{56.3} & 7,461 & 46.3 & 10,683 & 31.8 & 7,437 & 67.8 & 100\% \\
NoThinking & 84.6 & 231 & 79.4 & 970 & 73.1 & 1,807 & 41.0 & 1,998 & 22.5 & 5,551 & 28.8 & 638 & 54.9 & \textbf{30.6\%} \\
Dynasor & 89.8 & 773 & 89.4 & 2,987 & 86.7 & 4,689 & 54.8 & 6,477 & 42.5 & 9,726 & 31.3 & 1,471 & 65.8 & 71.4\% \\
DEER & 89.2 & 686 & 90.4 & 2,157 & 86.9 & 4,220 & 54.4 & 5,614 & 45.0 & 9,374 & 34.3 & 6,586 & 66.7 & 78.3\% \\
\rowcolor[HTML]{D5E8D4}
SABER \textit{(Ours)} & 91.0 & 662 & \textbf{91.2} & 2,401 & 90.0 & 4,256 & 56.1 & 5,560 & \textbf{50.9} & 9,333 & \textbf{34.8} & 3,326 & \textbf{69.0} & 69.8\% \\
\addlinespace[1pt]
% 模型区块 2
\specialrule{0.05pt}{1pt}{1pt}
\rowcolor{gray!8}
\multicolumn{15}{l}{\textit{\textbf{Qwen3-4B}}} \\
Vanilla & 94.7 & 2,113 & 92.4 & 4,796 & \textbf{92.5} & 7,396 & 59.7 & 9,066 & \textbf{56.3} & 12,399 & 53.5 & 8,139 & 74.8 & 100\% \\
NoThinking & 92.1 & 369 & 80.0 & 2,079 & 85.0 & 4,033 & 52.2 & 4,472 & 43.8 & 7,982 & 38.9 & 565 & 65.3 &\textbf{ 44.4\%} \\
Dynasor & 94.2 & 1,998 & 89.0 & 4,189 & 83.1 & 6,457 & 58.5 & 7,658 & 52.1 & 10,648 & 54.6 & 6,687 & 71.9 & 85.7\% \\
DEER & 94.3 & 1,204 & 91.4 & 3,420 & 89.4 & 5,714 & 60.1 & 6,680 & 52.1 & 9,706 & 53.5 & 7,012 & 73.5 & 76.8\% \\
\rowcolor[HTML]{D5E8D4}
SABER \textit{(Ours)} & \textbf{94.8} & 1,389 &\textbf{92.8} & 2,743 & 90.6 & 4,824 & \textbf{61.9} & 5,270 & 54.2 & 9,327 & \textbf{56.1} & 6,894 & \textbf{75.1} & 69.3\% \\
\addlinespace[1pt]
% 模型区块 3
\specialrule{0.05pt}{1pt}{1pt}
\rowcolor{gray!8}
\multicolumn{15}{l}{\textit{\textbf{Qwen3-8B}}} \\
Vanilla & \textbf{96.4} & 2,192 & \textbf{92.8} & 5,135 & \textbf{90.6} & 7,871 & 59.0 & 9,447 & \textbf{60.5} & 12,501 & 55.1 & 9,047 & 75.7 & 100\% \\
NoThinking & 91.3 & 383 & 85.8 & 2,372 & 86.3 & 4,692 & 51.9 & 4,939 & 43.8 & 7,907 & 50.5 & 3,423 & 68.3 & \textbf{51.3\%} \\
Dynasor & 95.1 & 1,740 & 89.8 & 4,060 & 86.9 & 6,225 & 57.9 & 7,415 & 52.5 & 10,475 & 52.0 & 4,420 & 72.4 & 74.3\% \\
DEER & 95.3 & 1,069 & 91.6 & 2,958 & 89.4 & 5,630 & 59.6 & 6,674 & 54.9 & 9,547 & 57.6 & 8,002 & 74.7 & 73.3\% \\
\rowcolor[HTML]{D5E8D4}
SABER \textit{(Ours)} & 96.0 & 1,258 & 92.4 & 2,303 & 90.0 & 4,173 & \textbf{61.5} & 4,892 & 58.8 & 10,119 & \textbf{57.6} & 5,056 &\textbf{76.0}&60.2\% \\
\bottomrule
\end{tabular}
\caption{
Experimental results across multiple reasoning benchmarks and model scales. ``Acc'' denotes accuracy, ``Tok'' denotes reasoning token count, and ``CR'' denotes compression ratio.$\uparrow$($\downarrow$) indicates higher (lower) values are better.The best results are highlighted in \textbf{bold}.
}
\label{tab:main_results}
\end{table*}
\paragraph{Benchmarks and Metrics.} 
We evaluate our method on six widely used mathematical reasoning benchmarks---GSM8K~\cite{cobbe2021trainingverifierssolvemath}, MATH-500~\cite{hendrycks2021measuringmathematicalproblemsolving}, AMC23~\cite{aimo_validation_amc}, OlympiadBench~\cite{olympiadbench}, AIME 2024, and AIME 2025~\cite{aime_aops}---as well as the scientific reasoning benchmark GPQA Diamond~\cite{gpqa}. We report three primary evaluation metrics: \textit{Accuracy} (\textbf{Acc}), which measures final answer correctness; \textit{Token Count} (\textbf{Tok}), representing the average number of generated tokens per sample; and \textit{Compression Ratio} (\textbf{CR}), which measures token consumption relative to vanilla reasoning, indicates inference efficiency, where lower values correspond to better performance.

\paragraph{Models.} 
We conduct experiments on a diverse set of reasoning LLMs spanning multiple scales and architectures, including \textbf{Qwen3-4B}, \textbf{Qwen3-8B}~\cite{yang2025qwen3}, and \textbf{DeepSeek-R1-Distilled-Qwen-7B}~\cite{guo2025deepseek}. For each model, we adopt the recommended reasoning prompts and decoding configurations.

\paragraph{Baselines.} 
We compare \textbf{SABER} against vanilla reasoning and several representative training-free efficient reasoning methods, including NoThinking~\cite{nothinking}, DEER~\cite{yang2025dynamic}, and Dynasor~\cite{fu2025reasoning}. \textbf{Vanilla reasoning} generates complete reasoning trajectories without early termination. \textbf{NoThinking} directly produces final answers, bypassing explicit reasoning. \textbf{DEER} performs adaptive early exit using confidence-based stopping signals, while \textbf{Dynasor} performs adaptive early exit based on intermediate answer consistency estimated through probing.

\paragraph{Implementation Details.} 
% All experiments are conducted using the vLLM framework. Unless otherwise specified, the temperature and top-$p$ values are set to 0.6 and 0.95, respectively. Due to the relatively small sizes of AMC 2023, AIME 2024, and AIME 2025, each instance is evaluated four times and results are averaged across runs. All experiments are conducted on a server equipped with 8 NVIDIA RTX 3090 GPUs. Additional implementation details and hyperparameter settings are provided in Appendix~\ref{appendix:hyperparameter}.
All experiments are conducted using the vLLM framework on a server equipped with 8 NVIDIA RTX 3090 GPUs. Due to the relatively small sizes of AMC 2023, AIME 2024, and AIME 2025, each instance is evaluated four times and results are averaged across runs. The sensitivity coefficient $\gamma$ in the confidence stability score is set to 3 across all experiments. The early-exit threshold $\tau$ is set to 0.9 for DeepSeek-R1-Distilled-Qwen-7B and 0.95 for the Qwen3-series models. Additional implementation details and hyperparameter settings are provided in Appendix~\ref{appendix:hyperparameter}.

\subsection{Main Results}
\label{subsec:main_results}

\paragraph{Overall Performance.} 
Table~\ref{tab:main_results} shows that SABER consistently achieves strong efficiency--accuracy trade-offs across different model scales and reasoning benchmarks. Compared with vanilla reasoning, SABER reduces reasoning tokens by 30.2\%--39.8\% while maintaining comparable or improved accuracy. In particular, SABER improves the overall accuracy of DeepSeek-R1-Distilled-Qwen-7B from 67.8 to 69.0 while using only 69.8\% of the original reasoning tokens, and also achieves the best overall accuracy on both Qwen3-4B and Qwen3-8B. These results suggest that our stability-aware early exit approach can effectively reduce redundant reasoning while preserving quality.

\paragraph{Comparison with Baselines.}
Compared with representative training-free efficient reasoning baselines, SABER consistently achieves stronger performance across diverse benchmarks and model scales. NoThinking attains high compression ratios by bypassing explicit reasoning, but suffers substantial accuracy degradation. Dynasor and DEER rely on prediction consistency and confidence-based early exiting, respectively, yet both often produce unreliable exit decisions when intermediate reasoning states remain uncertain. In contrast, SABER jointly integrates semantic consistency and confidence stability under adversarial perturbations, enabling more reliable reasoning convergence estimation. The gains are particularly evident on challenging benchmarks such as OlympiadBench and GPQA-D.
\section{Analysis}
\label{sec:Analysis}

\begin{figure}[t]
    \centering
    \includegraphics[width=\columnwidth]{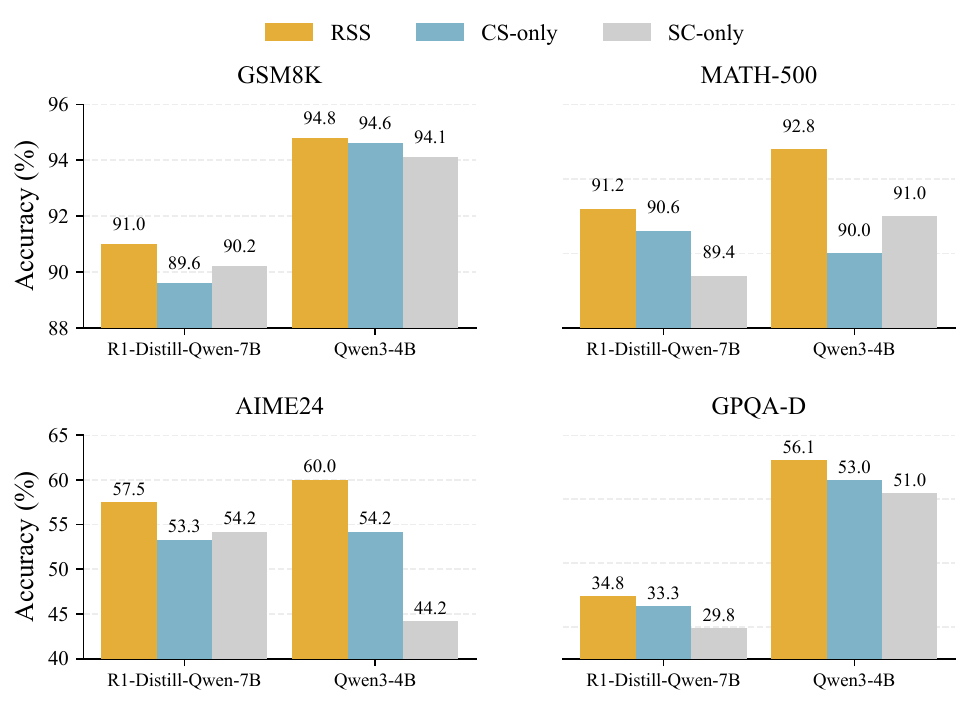}
    \caption{Accuracy of RSS early-exit on mathematical reasoning benchmarks: component ablation of semantic consistency (SC) and confidence stability (CS). Results are shown for RSS (our full method) compared with SC-only and CS-only variants on two representative models.}
    \label{fig:ablation}
\end{figure}

\begin{figure}[t]
    \centering
    \includegraphics[width=\columnwidth]{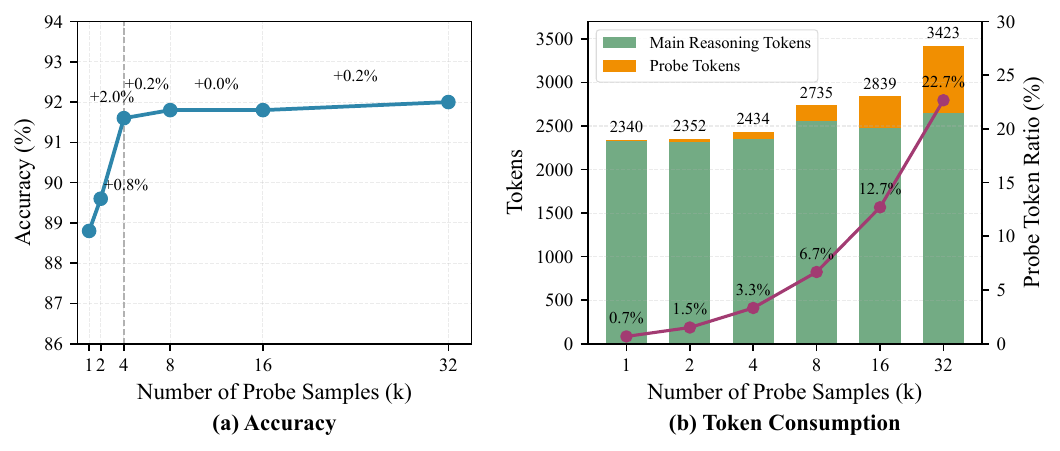}
    \caption{Sampling number ablation on MATH-500 using Qwen3-8B. Left: Accuracy improvement with increasing $k$. Right: Probe overhead growth and its proportion of total generated tokens as a function of $k$.}
    \label{fig:sampling_ablation}
\end{figure}

\begin{table}[t]
\centering
\small
\setlength{\tabcolsep}{3.5pt}
\begin{tabular}{l c c c}
\toprule
Model & $\alpha=0.3$ & $\alpha=0.5$ & $\alpha=0.7$ \\
\midrule
\multicolumn{4}{l}{\textbf{GSM8K}} \\
R1-Distill-Qwen-7B & 90.4/739 & 90.1/652 & \textbf{91.0}/662 \\
Qwen3-4B & 94.3/1,635 & 94.5/1,503 & \textbf{94.8}/1,389 \\
Qwen3-8B & 95.3/1,604 & 95.1/1,394 & \textbf{96.0}/1,258 \\
\midrule
\multicolumn{4}{l}{\textbf{OlympiadBench}} \\
R1-Distill-Qwen-7B & \textbf{56.1}/5,560 & 54.4/5,462 & 53.8/5,122 \\
Qwen3-4B & \textbf{61.9}/5,270 & 59.6/5,121 & 59.6/4,889 \\
Qwen3-8B & \textbf{61.5}/4,892 & 58.8/4,330 & 58.4/4,316 \\
\bottomrule
\end{tabular}
\caption{Sensitivity analysis of weight $\alpha$ on GSM8K and OlympiadBench. Each cell shows accuracy (\%) and average token count.}
\label{tab:alpha_sensitivity}
\end{table}

% \subsection{Ablation Study}
% To isolate the contribution of each signal in SABER, we conduct component ablation experiments comparing three configurations: (1) \textbf{SC-only}, using only semantic consistency as the early-exit signal, (2) \textbf{CS-only}, using only confidence stability as the early-exit signal, and (3) \textbf{RSS}, our full method. Figure~\ref{fig:ablation} shows the accuracy of two reasoning models across four benchmarks.

% Overall, RSS outperforms both single-signal variants across all tasks, highlighting the importance of jointly modeling semantic consistency and confidence stability. On relatively simple tasks such as GSM8K and MATH-500, all methods achieve comparable performance. However, on more challenging reasoning tasks such as AIME24 and GPQA-D, both SC-only and CS-only degrade noticeably.
% Further analysis reveals the complementary nature of the two signals. SC-only may trigger premature exits on semantically consistent yet incorrect answers, whereas CS-only is vulnerable to stable but incorrect reasoning trajectories. By jointly considering semantic consistency and confidence stability, RSS more reliably identifies genuinely converged reasoning states.

\subsection{Ablation Study}

\textbf{Component Ablation.}
To examine the contributions of the two stability signals in SABER, we compare three configurations: (1) \textbf{SC-only}, using only semantic consistency as the early-exit signal, (2) \textbf{CS-only}, using only confidence stability as the early-exit signal, and (3) \textbf{RSS}, our full method. Figure~\ref{fig:ablation} shows the accuracy of two reasoning models across four benchmarks. Overall, RSS outperforms both single-signal variants across all tasks, highlighting the importance of jointly modeling semantic consistency and confidence stability. On relatively simple tasks such as GSM8K and MATH-500, all methods achieve comparable performance. However, on more challenging reasoning tasks such as AIME24 and GPQA-D, both SC-only and CS-only degrade noticeably. This reflects the complementary nature of the two signals: SC-only may trigger premature exits on semantically consistent yet incorrect answers, whereas CS-only is vulnerable to stable but incorrect reasoning trajectories. By jointly considering semantic consistency and confidence stability, RSS more reliably identifies genuinely converged reasoning states.

\textbf{Scoring Function Ablation.}
To disentangle the contribution of the two-branch probing framework from that of the specific scoring function, we keep the neutral/adversarial two-branch probing framework unchanged and replace only the stopping score with two alternatives: (1) \textbf{SC $\cdot$ CS}, which multiplicatively combines semantic consistency and confidence stability, and (2) \textbf{Branch-UQ Diff}, which estimates the uncertainty of each branch using answer-set entropy and perplexity, and uses their difference as the stopping signal. Detailed formulations and results are provided in Appendix~\ref{app:scoring_ablation}. Both alternatives remain competitive under the same probing framework, indicating that SABER's gains are not tied to the specific formulation of RSS and suggesting that the two-branch probing framework is the primary source of the gains. Nevertheless, RSS achieves better performance, demonstrating its effectiveness as a stopping criterion within this framework.

\subsection{Impact of Sampling Number}
The sampling number $k$ plays a critical role in balancing accuracy and computational overhead in SABER. We evaluate the effect of different sampling numbers on Qwen3-8B using six configurations with $k \in \{1,2,4,8,16,32\}$. Figure~\ref{fig:sampling_ablation} presents the trends of accuracy, main reasoning tokens, and probe overhead on MATH-500.
As $k$ increases, model accuracy generally improves, indicating that larger sample sizes yield more stable convergence estimation. Raising $k$ from 1 to 4 increases accuracy from 88.8\% to 91.6\%, showing that a small number of additional samples significantly enhances the reliability of early-exit decisions.
However, further increasing $k$ leads to rapidly diminishing returns. Beyond $k=4$, accuracy improves by only 0.4\% even when the sampling number is raised to 32. Meanwhile, probe overhead grows approximately linearly, with its proportion of total tokens increasing from 3.3\% to 22.7\%, resulting in higher computational cost.
These results suggest diminishing returns from increasing sampling budget, with most benefits achieved at small $k$ values.

\subsection{Effect of Weight \texorpdfstring{$\alpha$}{alpha}}
To investigate the impact of the balancing weight $\alpha$, we conduct a sensitivity analysis across different models and datasets, with results shown in Table~\ref{tab:alpha_sensitivity}. On GSM8K, $\alpha = 0.7$ consistently achieves the highest accuracy with fewer tokens across all three models, indicating that semantic consistency plays a dominant role in supporting reliable early exit on relatively simple reasoning tasks. In contrast, on OlympiadBench, smaller $\alpha$ values consistently yield higher accuracy, suggesting that difficult reasoning tasks rely more heavily on confidence stability, since semantic consistency alone may produce stable yet incorrect reasoning trajectories. Overall, simple tasks favor larger $\alpha$ values, while more challenging tasks benefit from smaller $\alpha$ values. These observations further demonstrate the complementarity of semantic consistency and confidence stability, and highlight a clear task-dependent preference in balancing the two signals across different levels of reasoning difficulty.

\begin{figure}[t]
    \centering
    \includegraphics[width=\columnwidth]{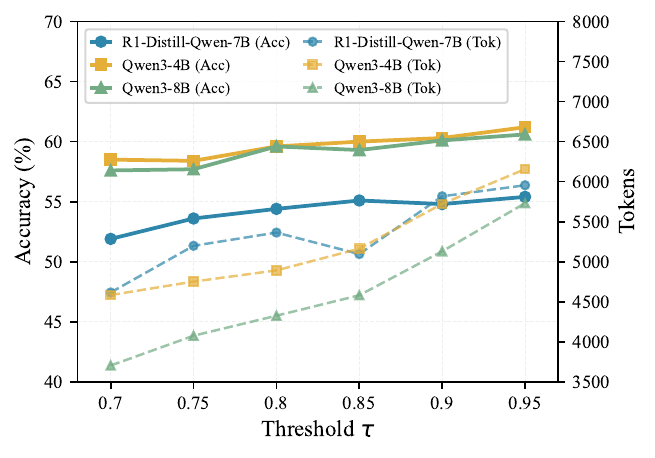}
    \caption{Accuracy (left y-axis) and average token count (right y-axis) on OlympiadBench under different stopping thresholds $\tau$ with $\alpha=0.3$. Results are shown for three reasoning models.}
    \label{fig:tau}
\end{figure}

\subsection{Threshold Sensitivity \texorpdfstring{$\tau$}{tau}}
We conduct threshold sensitivity analysis on the challenging OlympiadBench benchmark, since difficult reasoning tasks are generally more sensitive to early-exit decisions. Figure~\ref{fig:tau} shows the accuracy and average token count of three models under different stopping thresholds with $\alpha = 0.3$. As $\tau$ increases, the model becomes more selective, postponing early exit until stronger stability is achieved, thereby reducing premature exits on ambiguous reasoning steps. Consequently, model accuracy improves, accompanied by higher token consumption. Within $\tau \in [0.8, 0.95]$, performance remains relatively stable, indicating that SABER is robust to threshold selection within practical ranges and does not require delicate tuning.

\subsection{Probe Overhead Analysis}
% \begin{table}[t]
% \centering
% \small
% \begin{tabular}{l c c c}
% \toprule
% Model & Probe & Total & Ratio (\%) \\
% \midrule
% R1-Distill-Qwen-7B & 221 & 4,477 & 4.9\% \\
% Qwen3-4B & 147 & 5,221 & 2.8\% \\
% Qwen3-8B & 181 & 4,815 & 3.8\% \\
% \midrule
% Average & 183 & 4,838 & 3.8\% \\
% \bottomrule
% \end{tabular}
% \caption{
% Probe overhead under the default configuration for DeepSeek-R1-Distill-Qwen-7B, Qwen3-4B, and Qwen3-8B. Probe tokens account for only a small fraction of total generated tokens.
% }
% \label{tab:overhead}
% \end{table}

\begin{table}[t]
\centering
\small
\setlength{\tabcolsep}{8pt}

\begin{tabular}{lccc}
\toprule
Model & Probe & Total & Ratio (\%) \\
\midrule
R1-Distill-Qwen-7B & 221 & 4,477 & 4.9\% \\
Qwen3-4B & 147 & 5,221 & 2.8\% \\
Qwen3-8B & 181 & 4,815 & 3.8\% \\
\midrule
Average & 183 & 4,838 & 3.8\% \\
\bottomrule
\end{tabular}

\caption{
Probe overhead under the default configuration. Probe tokens account for only a small fraction of total generated tokens.
}
\label{tab:overhead}
\end{table}
We analyze the additional computational overhead introduced by the probing mechanism in SABER. 
As shown in Table~\ref{tab:overhead}, which reports the average probing ratios across all test datasets, the ratios for the three models are 4.9\%, 2.8\%, and 3.8\%, respectively, with an average of only 3.8\%. This indicates that adversarial probing constitutes a relatively small proportion of the overall reasoning process, and the associated overhead remains manageable.
More importantly, the token savings achieved through early exit substantially outweigh the additional cost introduced by adversarial probing, demonstrating a favorable cost-benefit trade-off.

\subsection{Inference Latency}
% \input{tables/latency_token}
% \begin{table}[t]
% \centering
% \small
% \begin{tabular}{l c c c}
% \toprule
% Model & Vanilla & SABER & Reduction \\
% \midrule
% R1-Distill-Qwen-7B & 149.4 & 92.9 & 37.8\% \\
% Qwen3-4B & 190.1 & 107.0 & 43.7\% \\
% Qwen3-8B & 243.0 & 98.5 & 59.5\% \\
% \midrule
% Average & 194.2 & 99.5 & 48.8\% \\
% \bottomrule
% \end{tabular}
% \caption{Average inference latency (in seconds) of SABER across GSM8K, AIME24, and GPQA-D.}
% \label{tab:latency_reduction}
% \end{table}

\begin{table}[t]
\centering
\small
\setlength{\tabcolsep}{6pt}

\begin{tabular}{lccc}
\toprule
Model & Vanilla & SABER & Reduction \\
\midrule
R1-Distill-Qwen-7B & 149.4 & 92.9 & 37.8\% \\
Qwen3-4B & 190.1 & 107.0 & 43.7\% \\
Qwen3-8B & 243.0 & 98.5 & 59.5\% \\
\midrule
Average & 194.2 & 99.5 & 48.8\% \\
\bottomrule
\end{tabular}

\caption{
Average inference latency (in seconds) of SABER across GSM8K, AIME24, and GPQA-D.
}
\label{tab:latency_reduction}
\end{table}
We evaluate the end-to-end inference latency of SABER on GSM8K, AIME24, and GPQA-D, covering reasoning tasks of varying difficulty levels. Table~\ref{tab:latency_reduction} reports the average inference latency across three models.
The results show that SABER significantly reduces inference latency (48.8\% on average). The reduction is particularly pronounced on more challenging tasks with longer reasoning trajectories, where late-stage computation dominates runtime.
These latency gains are consistent with our token reduction analysis, suggesting that fewer generated tokens contribute to lower wall-clock time.
In summary, SABER achieves substantial wall-clock speedups, highlighting its practical efficiency in real-world inference scenarios.

% \subsection{Perturbation Design Discussion}
% We define a perturbation family $\mathcal{P}$ to probe trajectory robustness. To evaluate perturbation quality, we introduce an indicator $G(p)$ that measures the discriminative power between correct and incorrect reasoning trajectories under perturbation $p$. Our goal is to identify the perturbation that maximizes this discriminative power. Implementation details are provided in Appendix~\ref{sec:gp}.
\section{Related Work}
\label{sec:related_work}

\paragraph{Efficient Reasoning and Early Exit.}
To mitigate the overthinking problem, researchers have proposed various efficient reasoning strategies, which can be broadly categorized into three types.
\textbf{Post-training methods} enable models to adaptively generate reasoning chains of varying lengths through supervised fine-tuning on variable-length CoT data~\citep{zhao2025let,yu2024distilling,kang2025c3ot} or length rewards in reinforcement learning~\citep{lou2025adacot,aggarwal2025l1}.
\textbf{Prompt-guided methods} rely on carefully designed prompts, such as "Be concise"~\citep{renze2024benefits,xu2025chain} or token budget constraints~\citep{han2025token}, to encourage shorter reasoning. 
\textbf{Inference-time methods} dynamically control reasoning behavior during generation to reduce redundant computation without modifying model parameters, offering greater flexibility. We focus on inference-time methods and study how to determine when reasoning has been sufficient for early termination.

\paragraph{Inference-time Early Exit Signals.}
The core of inference-time early exit methods lies in constructing a reliable \textbf{stopping signal}. 
One representative line of work is based on \textbf{uncertainty signals}, such as confidence or entropy. EAT~\citep{wang2025entropy} leverages token-level entropy, HALT-CoT~\citep{laaouach2025halt} computes the entropy of answer distributions, while DEER~\citep{yang2025dynamic} and CGRS~\citep{huang2026efficient} decide whether to exit at specific checkpoints based on confidence.
Another line of work relies on \textbf{consistency signals}. LRMs often converge to stable final answers at early stages, which can be directly used as a stopping signal~\citep{liu2025answer}. Dynasor~\citep{fu2025reasoning} triggers early exit by detecting consistency in intermediate answers, and TRACE~\citep{li2026efficient} further combines consistency and confidence within a sliding window, at the cost of multi-step aggregation and additional latency.
In addition, \textbf{hidden-state methods} attempt to estimate reasoning status directly from internal representations. TPV~\citep{eisenstadt2025overclocking} encodes reasoning progress using hidden states, SEAL~\citep{chen2025seal} steers reasoning via activation vectors, and probe-based methods~\citep{zhang2025reasoning} train probes on hidden states to predict answer correctness. However, these methods typically require additional training and model-specific adaptation.
Overall, existing methods primarily rely on output-level or internal signals for early exit, while leaving the stability of intermediate reasoning trajectories underexplored as a factor for decision making.

\paragraph{Multi-path Reasoning and Perturbation.}
To obtain more reliable predictions than a single reasoning chain, one line of research uses multi-path reasoning. Self-Consistency~\citep{wang2022self} generates multiple reasoning paths via repeated sampling and aggregates final predictions through voting, based on the insight that semantic agreement across paths reflects answer reliability.
However, this approach requires generating full-length paths in parallel and is not directly applicable to early exit.
A related line of research studies reasoning path perturbation, which injects neutral or adversarial perturbations into intermediate CoT steps~\citep{von2026reasoning,aravindan2026fragile}, revealing the fragility and limited recovery ability of reasoning models under perturbations. Meanwhile, prior studies have shown that behavioral stability under perturbations can provide useful signals for model reliability and uncertainty estimation~\citep{khanmohammadi2025ccps,saadat2026certainty}.
However, existing perturbation-based studies primarily use perturbations for robustness evaluation or behavioral analysis, rather than online reasoning control during generation.
Building on these insights, we view perturbation-induced multi-branch reasoning as a promising source of stability signals for early exit.

% \paragraph{Fundamental Limitation: Confident but Wrong}
% Although they take different forms, existing early exit methods share a common premise: \textbf{deterministic signals from the model (e.g., high confidence, low entropy, or high consistency) can serve as a proxy for correctness}. However, this assumption systematically fails in reasoning models.
% Research on confidence calibration has revealed that LLMs suffer from severe overconfidence. Furthermore, extending the reasoning process does not alleviate this issue; instead, it may \textbf{amplify confidence in incorrect intermediate conclusions} ~\citep{mei2026reasoning}. On challenging tasks, the average confidence for incorrect answers often exceeds 85\%. More critically, models can fall into a \textbf{"confidently wrong"} state: different reasoning paths converge on the same incorrect answer, and models may even exhibit higher confidence when hallucinating ~\citep{slobodkin2023curious,simhi2025trust,kadavath2022language}.
% This means that confidence- and entropy-based methods may prematurely exit with wrong answers, while consistency-based methods may mistakenly terminate due to erroneous convergence. Consequently, the limitation of these methods stems not from their specific metric design, but from the fact that the signals they rely on are inherently incapable of distinguishing between "confidently correct" and "confidently wrong."
\section{Conclusion}
\label{sec:Conclusion}
We propose SABER, a training-free framework for stability-aware early exit via adversarial branch probing. Rather than relying solely on confidence or entropy, SABER estimates reasoning convergence by measuring the stability of intermediate reasoning states under lightweight semantic perturbations. This approach efficiently evaluates the consistency of potential outcomes without requiring additional full-generation rollouts. Across diverse reasoning benchmarks and model architectures, SABER substantially reduces reasoning token consumption while maintaining or even improving reasoning accuracy. Overall, our results suggest that reasoning stability provides an effective signal for adaptive inference in large reasoning models and offers a promising direction for efficient inference-time scaling. 
\section*{Limitations}
\label{sec:Limitations}
Despite the promising results of SABER in improving reasoning efficiency, our evaluation is subject to several limitations. Due to computational constraints, we conduct experiments on text-based reasoning benchmarks using models ranging from 4B to 8B parameters, and do not investigate larger-scale models. In addition, we focus exclusively on text-based settings and do not evaluate SABER in multimodal or agent-based scenarios, where intermediate representations and confidence dynamics may differ significantly. Exploring the applicability of SABER in these broader settings remains an important direction for future work.

\section*{Acknowledgments}
We would like to thank all the anonymous reviewers for their valuable comments. This work was supported by the National Natural Science Foundation of China (NSFC No. 62576232) and the Key Laboratory of Data Intelligence and Advanced Computing, Soochow University.

% This document has been adapted
% by Steven Bethard, Ryan Cotterell and Rui Yan
% from the instructions for earlier ACL and NAACL proceedings, including those for
% ACL 2019 by Douwe Kiela and Ivan Vuli\'{c},
% NAACL 2019 by Stephanie Lukin and Alla Roskovskaya,
% ACL 2018 by Shay Cohen, Kevin Gimpel, and Wei Lu,
% NAACL 2018 by Margaret Mitchell and Stephanie Lukin,
% Bib\TeX{} suggestions for (NA)ACL 2017/2018 from Jason Eisner,
% ACL 2017 by Dan Gildea and Min-Yen Kan,
% NAACL 2017 by Margaret Mitchell,
% ACL 2012 by Maggie Li and Michael White,
% ACL 2010 by Jing-Shin Chang and Philipp Koehn,
% ACL 2008 by Johanna D. Moore, Simone Teufel, James Allan, and Sadaoki Furui,
% ACL 2005 by Hwee Tou Ng and Kemal Oflazer,
% ACL 2002 by Eugene Charniak and Dekang Lin,
% and earlier ACL and EACL formats written by several people, including
% John Chen, Henry S. Thompson and Donald Walker.
% Additional elements were taken from the formatting instructions of the \emph{International Joint Conference on Artificial Intelligence} and the \emph{Conference on Computer Vision and Pattern Recognition}.

\bibliography{custom}

\appendix
\section{Experimental Details}
\label{appendix:experimental}

\subsection{Hyperparameter Settings}
\label{appendix:hyperparameter}
We provide detailed hyperparameter settings of SABER, including configurations for adversarial probing, stability estimation, and early-exit decisions. During adversarial probing, we sample short continuations to estimate reasoning trajectory stability under perturbations. To reduce the additional computational overhead introduced by probing, each probing branch is restricted to generate at most 10 tokens. At each intermediate reasoning step, we perform stochastic decoding with sampling number $k=4$ to obtain sampled answer sets and corresponding confidence sets, as defined in Eq.~\ref{eq:ans} and Eq.~\ref{eq:confidence}. Unless otherwise specified, we use a temperature of 0.6 and a top-$p$ of 0.95.

For all experiments, the maximum context length and maximum generation length are set to 32k and 16k tokens, respectively. For the confidence stability score in Eq.~\ref{eq:cs}, we set the sensitivity coefficient $\gamma=3$ across all experiments to better capture confidence fluctuations under adversarial perturbations. For the RSS in Eq.~\ref{eq:rss}, we perform hyperparameter search over $\alpha \in \{0.3, 0.5, 0.7\}$, with 0.3 or 0.7 selected depending on the model and dataset. Finally, we perform hyperparameter search over $\tau \in \{0.85, 0.9, 0.93, 0.95, 0.98\}$, with the selected threshold set to 0.9 for DeepSeek-R1-Distilled-Qwen-7B and 0.95 for the Qwen3-series models.

\subsection{Dataset Details}

The mathematical and scientific reasoning datasets used in our evaluation are described below. We strictly follow the licenses and evaluation protocols specified in the original papers.

\textbf{GSM8K}~\cite{cobbe2021trainingverifierssolvemath} is a widely used benchmark for grade-school mathematical reasoning. It consists of diverse multi-step arithmetic word problems that require sequential reasoning and intermediate numerical calculations, making it a standard benchmark for evaluating chain-of-thought reasoning capability.

\textbf{MATH-500}~\cite{hendrycks2021measuringmathematicalproblemsolving} is a curated subset of the MATH benchmark containing 500 high-difficulty mathematics problems. The dataset spans multiple domains, including algebra, calculus, geometry, and probability, and is widely used for evaluating mathematical reasoning and test-time scaling behavior.

\textbf{AMC23}~\cite{aimo_validation_amc} is a validation benchmark constructed from AMC 12 problems released in 2023. The dataset focuses on high-school-level mathematics and includes problems involving algebraic reasoning, geometry, and combinatorial counting. Following prior work, all answers are normalized to integer outputs for consistent evaluation.

\textbf{OlympiadBench}~\cite{olympiadbench} is a challenging benchmark designed to evaluate advanced mathematical reasoning at the Olympiad level. It consists of competition-style problems covering algebra, geometry, number theory, and combinatorics. The benchmark typically requires long-form multi-step reasoning and symbolic manipulation, making it particularly challenging for large reasoning models.

\textbf{AIME 2024} and \textbf{AIME 2025}~\cite{aime_aops} are constructed from the American Invitational Mathematics Examination problems from 2024 and 2025, respectively. These datasets consist of short-answer competition math problems that require deep multi-step reasoning and nontrivial mathematical insights.

\textbf{GPQA Diamond}~\cite{gpqa} is the difficult split of the Graduate-Level Google-Proof QA benchmark. It contains expert-written multiple-choice questions spanning physics, chemistry, and biology, where solving the problems requires domain knowledge and multi-step scientific reasoning rather than memorization.

All datasets are evaluated under the zero-shot inference setting. Their requirement for long reasoning chains and complex intermediate computations makes them well-suited for studying early-exit strategies and reasoning efficiency.

\subsection{Prompt Templates}
We use a unified prompting format across all benchmarks to ensure fair comparison between different early-exit methods. For all datasets, the model is instructed to explicitly generate intermediate reasoning steps before producing the final answer.

\textbf{Base Inference Prompt.}
The following prompt is used for standard reasoning generation during inference:
\begin{promptbox}
\texttt{Please reason step by step, and put your final answer within \textbackslash boxed\{\}.}
\end{promptbox}

\textbf{Neutral Probing Prompt.}
The neutral probing prompt encourages the model to continue its current reasoning trajectory and summarize the inferred answer:

\begin{promptbox}
\texttt{Wait, let me summarize. The answer is \textbackslash boxed\{}
\end{promptbox}

\textbf{Adversarial Probing Prompt.}
The adversarial probing prompt encourages the model to suspect that its previous reasoning trajectory may be incorrect, and then guides the model to regenerate and summarize the final answer under the perturbed reasoning context:
\begin{promptbox}
\texttt{Wait, I think my previous reasoning was incorrect. After correcting it, the answer is \textbackslash boxed\{}
\end{promptbox}

\section{Implementation Details}
\label{appendix:implementation}

\subsection{Overview}
SABER is implemented as a training-free decoding-time intervention that operates seamlessly on top of the standard auto-regressive generation process. During inference, the framework continuously monitors the reasoning trajectory for predefined transition points. Once triggered, the current reasoning state is branched into lightweight neutral and adversarial probing continuations, where stochastic multi-sampling is performed in parallel. Based on the resulting Reasoning Stability Score (RSS), SABER dynamically determines whether reasoning has sufficiently converged: if the RSS exceeds a predefined threshold $\tau$, the framework injects the closure token \texttt{</think>} to terminate reasoning early and directly generate the final answer; otherwise, the probing branches are discarded and standard step-by-step reasoning resumes. To ensure practical efficiency, all probing branches are strictly length-constrained and parallelized within the inference engine.
\subsection{Pseudo-code}
Algorithm~\ref{alg:saber} summarizes the SABER procedure.
\begin{algorithm*}[t]
\caption{SABER: Stability-Aware Early Exit for LLM Reasoning via Adversarial Branch Probing}
\label{alg:saber}
\begin{algorithmic}[1]
\Require Query $q$; model $\mathcal{M}$; transition tokens $\mathcal{T}$ (e.g., ``Wait'', ``Alternatively''); sample size $k$; sensitivity $\gamma$; weight $\alpha$; threshold $\tau$; max length $L_{\max}$
\Ensure Final generated answer $y$
\State $\mathcal{P}_{\text{probe\_n}} \leftarrow \text{``Wait, let me summarize. The answer is \textbackslash boxed\{''}$ \Comment{Define neutral probe template}
\State $\mathcal{P}_{\text{probe\_a}} \leftarrow \text{``Wait, I think my previous reasoning was incorrect...''}$ \Comment{Define adversarial probe template}
\State $\mathcal{P} \leftarrow q$ \Comment{Initialize reasoning prefix state $\mathcal{P}$}
\While{$|\mathcal{P}| < L_{\max}$}
    \State Generate next token $t \sim \mathcal{M}(\mathcal{P})$ \Comment{Auto-regressive generation}
    \State $\mathcal{P} \leftarrow \mathcal{P} \oplus t$
    
    \If{$t \in \mathcal{T}$} \Comment{Intercept at transition points}
        \State $\mathcal{P}_n \leftarrow \mathcal{P} \oplus \mathcal{P}_{\text{probe\_n}}$ \Comment{Neutral probe prefix}
        \State $\mathcal{P}_a \leftarrow \mathcal{P} \oplus \mathcal{P}_{\text{probe\_a}}$ \Comment{Adversarial probe prefix}
        \State $\mathcal{A}_n, \mathcal{A}_a \leftarrow \emptyset$ \Comment{Initialize answer multisets, Eq.~(\ref{eq:ans})}
        \State $\mathcal{C}_n, \mathcal{C}_a \leftarrow \emptyset$ \Comment{Initialize confidence sets, Eq.~(\ref{eq:confidence})}
        
        \For{$j = 1$ to $k$}
            \State Sample $a_n^{(j)} \sim \mathcal{M}(\mathcal{P}_n)$; compute confidence $c_n^{(j)}$ 
            \State Sample $a_a^{(j)} \sim \mathcal{M}(\mathcal{P}_a)$; compute confidence $c_a^{(j)}$ 
            \State $\mathcal{A}_n \leftarrow \mathcal{A}_n \cup \{a_n^{(j)}\}$, \ $\mathcal{C}_n \leftarrow \mathcal{C}_n \cup \{c_n^{(j)}\}$
            \State $\mathcal{A}_a \leftarrow \mathcal{A}_a \cup \{a_a^{(j)}\}$, \ $\mathcal{C}_a \leftarrow \mathcal{C}_a \cup \{c_a^{(j)}\}$
        \EndFor
        
        \State $SC \leftarrow \textsc{MultisetJaccard}(\mathcal{A}_n, \mathcal{A}_a)$ \Comment{Semantic Consistency, Eq.~(\ref{eq:sc})}
        \State $\bar{P}_n \leftarrow \frac{1}{k}\sum_{j=1}^{k} c_n^{(j)}, \quad
        \bar{P}_a \leftarrow \frac{1}{k}\sum_{j=1}^{k} c_a^{(j)}$ \Comment{Average confidences}
        \State $CS \leftarrow e^{-\gamma \cdot |\bar{P}_n - \bar{P}_a|}$ \Comment{Confidence Stability, Eq.~(\ref{eq:cs})}
        \State $RSS \leftarrow \alpha \cdot SC + (1 - \alpha) \cdot CS$ \Comment{Reasoning Stability Score, Eq.~(\ref{eq:rss})}
        
        \If{$RSS > \tau$}
            \State $\mathcal{P} \leftarrow \mathcal{P} \oplus \text{\texttt{</think>}}$ \Comment{Inject closure token}
            \State \textbf{break} \Comment{Early exit: reasoning state has converged}
        \EndIf
    \EndIf
\EndWhile
\State $y \sim \mathcal{M}(\mathcal{P})$ \Comment{Trigger final answer generation}
\State \Return $y$
\end{algorithmic}
\end{algorithm*}

\subsection{Reasoning Transition Points}
\label{appendix:rtp}
As discussed in Section~\ref{sec:motivation}, we segment reasoning trajectories into intermediate steps using several transition trigger phrases, including ``Wait'', ``Alternatively'', ``But'', ``So'', and ``Let me double-check''.
These phrases commonly indicate self-correction, verification, or subgoal transitions, and empirically provide effective decomposition of long reasoning trajectories into coherent reasoning steps.

In subsequent evaluations, to reduce probing overhead, we only perform adversarial perturbation probing when the token ``Wait'' appears, as it is the most representative indicator of reasoning revision and uncertainty reflection.

\section{Additional Experimental Results}
\label{appendix:results}
\begin{figure}[t]
    \centering
    \includegraphics[width=\linewidth]{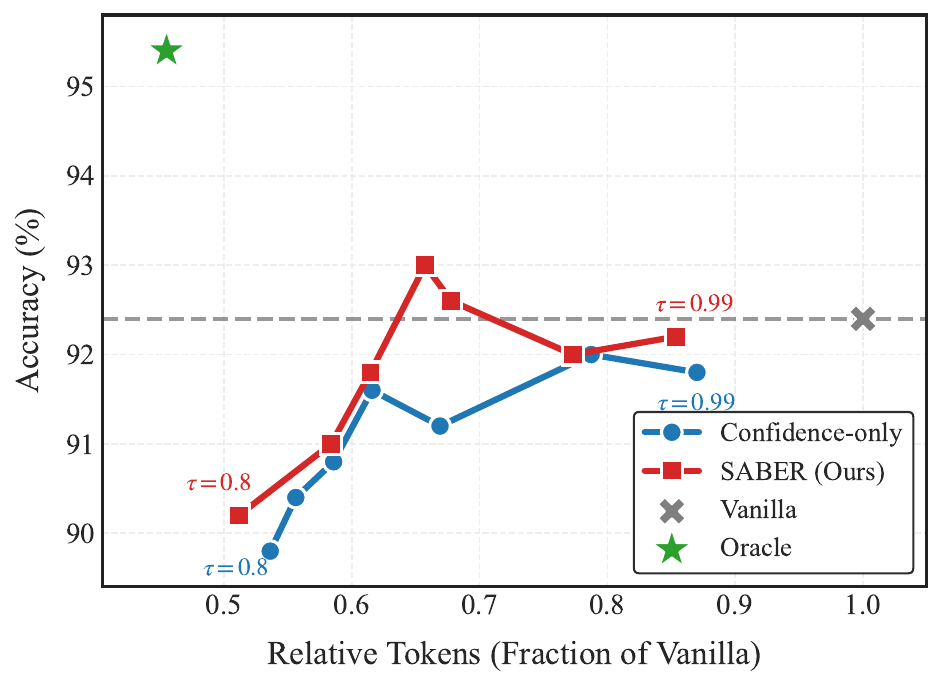}
    \caption{
    Comparison between SABER and a confidence-only early-exit strategy on MATH-500. SABER consistently achieves a better efficiency--accuracy trade-off and remains closer to the oracle upper bound across different stopping thresholds.
    }
    \label{fig:confidence_only}
\end{figure}

\subsection{Comparison with Confidence-Only Early-Exit}
\label{sec:confidence_only}
To isolate the contribution of adversarial branch probing, we compare SABER with a confidence-only early-exit variant that removes the adversarial branch and performs stopping decisions solely based on confidence estimated from the neutral branch.

Figure~\ref{fig:confidence_only} shows the efficiency--accuracy trade-off on MATH-500 under different stopping thresholds. SABER consistently achieves a better Pareto frontier across nearly all operating regions and remains substantially closer to the oracle upper bound.

These results suggest that confidence alone is insufficient for reliable early exit, since reasoning models may remain highly confident even when intermediate reasoning states have not fully converged. By evaluating reasoning stability under adversarial perturbations through semantic consistency and confidence stability, SABER provides more reliable reasoning stability estimation. Notably, SABER surpasses Vanilla reasoning under several thresholds while using fewer reasoning tokens.

\subsection{Scoring Function Ablation}
\label{app:scoring_ablation}
\begin{table*}[t]
\centering
\setlength{\tabcolsep}{3pt}
\begin{tabular}{l cc cc cc cc cc | cc}
\toprule
\multirow{2}{*}{\textbf{Scoring Function}}
& \multicolumn{2}{c}{\textbf{GSM8K}}
& \multicolumn{2}{c}{\textbf{MATH-500}}
& \multicolumn{2}{c}{\textbf{Olympiad}}
& \multicolumn{2}{c}{\textbf{AIME24,25}}
& \multicolumn{2}{c}{\textbf{GPQA-D}}
& \multicolumn{2}{c}{\textbf{Overall}} \\
& Acc$\uparrow$ & Tok$\downarrow$
& Acc$\uparrow$ & Tok$\downarrow$
& Acc$\uparrow$ & Tok$\downarrow$
& Acc$\uparrow$ & Tok$\downarrow$
& Acc$\uparrow$ & Tok$\downarrow$
& Acc$\uparrow$ & CR$\downarrow$ \\
\midrule
Vanilla
& \textbf{91.2} & 1,164
& 90.4 & 3,797
& \textbf{56.3} & 7,461
& 46.3 & 10,683
& 31.8 & 7,437
& 63.2 & 100\% \\

RSS (Ours)
& 91.0 & 662
& 91.2 & \textbf{2,401}
& 56.1 & 5,560
& \textbf{50.9} & 9,333
& \textbf{34.8} & 3,326
& \textbf{64.8} & 69.7\% \\

SC $\cdot$ CS
& 91.1 & \textbf{659}
& \textbf{91.4} & 2,509
& 56.1 & \textbf{5,505}
& 48.4 & 9,697
& 31.8 & 4,576
& 63.8 & 75.1\% \\

Branch-UQ Diff
& 90.1 & 864
& 90.6 & 2,826
& 55.0 & 5,639
& 47.1 & \textbf{8,788}
& 31.3 & \textbf{2,998}
& 62.8 & \textbf{69.1\%} \\
\bottomrule
\end{tabular}
\caption{Ablation of different scoring functions under the same neutral/adversarial probing framework on DeepSeek-R1-Distill-Qwen-7B.}
\label{tab:scoring_ablation}
\end{table*}
We provide detailed formulations and results for the alternative scoring functions evaluated under the same neutral/adversarial two-branch probing framework.

For \textbf{SC $\cdot$ CS}, we replace the weighted-additive formulation of RSS with a multiplicative combination of the same two stability signals:
\begin{equation}
S_{mult} = SC \cdot CS.
\end{equation}

% For \textbf{Branch-UQ Diff}, we instead adopt a branch-level uncertainty-based score. Specifically, we estimate the uncertainty of each branch $b \in \{n,a\}$, corresponding to the neutral and adversarial branches, as
% For \textbf{Branch-UQ Diff}, we instead adopt a branch-level uncertainty-based score. Specifically, we estimate the uncertainty of each branch as
For \textbf{Branch-UQ Diff}, we instead adopt a branch-level uncertainty-based score. Inspired by prior uncertainty quantification methods~\citep{vashurin2025uncertaintyquantificationllmsminimum}, we estimate the uncertainty of each branch as
\begin{equation}
UQ_b = H_b + \log PPL_b,
\end{equation}
where $H_b$ denotes the answer-set entropy and $PPL_b$ denotes the perplexity of branch $b$. The stopping score is then defined as the absolute uncertainty difference between the two branches:
\begin{equation}
S_{UQ} = |UQ_n - UQ_a|.
\end{equation}

For a fair comparison, we independently tune the stopping threshold for each alternative: $\tau \in \{0.75, 0.80, 0.85, 0.90, 0.95\}$ for SC $\cdot$ CS and $\tau \in \{0.05, 0.10, 0.20, 0.30\}$ for Branch-UQ Diff.

Table~\ref{tab:scoring_ablation} reports the results. Both SC $\cdot$ CS and Branch-UQ Diff remain effective under the same two-branch probing framework, despite using different scoring formulations. RSS achieves better performance because it additionally models semantic consistency between the two branches, rather than relying only on their uncertainty gap.

\subsection{Adversarial Probing Prompt Style Analysis}
\begin{table*}[t]
    \centering
    \begin{tabular}{p{3.4cm} p{10cm}}
        \toprule
        \textbf{Category} & \textbf{Perturbation Prompt} \\
        \midrule
        \multirow{2}{*}{Self-Correction} & ``Wait, I might have made a mistake. Let me reconsider, the answer is \texttt{\textbackslash boxed\{}'' \\
        \midrule
        Reflection & ``Wait, let me reconsider the reasoning, the answer is \texttt{\textbackslash boxed\{}'' \\
        \midrule
        Alternative Path & ``Wait, let me try a different approach, the answer is \texttt{\textbackslash boxed\{}'' \\
        \midrule
        \multirow{2}{*}{Verification} & ``Wait, let me verify the reasoning once more, the answer is \texttt{\textbackslash boxed\{}'' \\
        \bottomrule
    \end{tabular}
    \caption{Example perturbation probes. Each category represents a family of possible prompts; only one example is shown per category. SABER defaults to the Self-Correction style (not separately marked).}
    \label{tab:probe_family}
\end{table*}
\begin{table*}[t]
\centering
\setlength{\tabcolsep}{4pt}
\begin{tabular}{l cc cc cc cc | cc}
\toprule
\multirow{2}{*}{\textbf{Perturbation Strategy}}  
& \multicolumn{2}{c}{\textbf{GSM8K}} 
& \multicolumn{2}{c}{\textbf{MATH-500}} 
& \multicolumn{2}{c}{\textbf{Olympiad}} 
& \multicolumn{2}{c}{\textbf{GPQA-D}} 
& \multicolumn{2}{c}{\textbf{Overall}} \\
& Acc$\uparrow$ & Tok$\downarrow$ 
& Acc$\uparrow$ & Tok$\downarrow$ 
& Acc$\uparrow$ & Tok$\downarrow$ 
& Acc$\uparrow$ & Tok$\downarrow$ 
& Acc$\uparrow$ & Tok$\downarrow$ \\
\midrule
SABER (Ours)
& \textbf{94.4} & 1,349 
& \textbf{92.0} & 2,744 
& 60.0 & 5,918  
& \textbf{57.6} & 6,502
& \textbf{76.0} & 4,128 \\
Self-Correction
& 94.1 & 1,098 
& 91.6 & 2,591 
& 59.4 & 5,837 
& 57.0 & 5,497
& 75.5 & 3,756 \\
Reflection
& 94.2 & 1,085 
& 90.8 & 2,647 
& 58.5 & 5,600 
& 57.6 & \textbf{5,279}
& 75.3 & 3,653 \\
Alternative Path
& 94.1 & 1,173 
& 90.2 & 2,765 
& \textbf{61.0} & 5,875 
& 52.5 & 6,029
& 74.5 & 3,961 \\
Verification
& 94.2 & \textbf{1,023} 
& 91.6 & \textbf{2,538}
& 58.7 & \textbf{5,198} 
& 52.0 & 5,315
& 74.1 & \textbf{3,519} \\
\bottomrule
\end{tabular}
\caption{Comparison of different perturbation styles on four benchmarks using Qwen3-4B.}
\label{tab:prompt_ablation}
\end{table*}
To systematically study the effect of perturbation behaviors, we design a perturbation prompt family consisting of four intervention styles: \textit{Self-Correction} (explicitly acknowledging and correcting a potential mistake), \textit{Reflection} (reconsidering the current reasoning state), \textit{Alternative Path} (encouraging an alternative reasoning trajectory), and \textit{Verification} (checking the existing reasoning process), as shown in Table~\ref{tab:probe_family}. SABER uses a \textit{Self-Correction} style probe as the default configuration, whereas the \textit{Self-Correction} variant adopts an alternative prompt from the same perturbation family.

We then evaluate how different perturbation styles affect the behavior of SABER. Table~\ref{tab:prompt_ablation} compares our default SABER configuration against four perturbation variants using Qwen3-4B. 
Among the four variants, \textit{Self-Correction} achieves the highest average accuracy with moderate token consumption, while \textit{Verification} is the most token-efficient with competitive accuracy. Notably, our default SABER configuration consistently outperforms all variants, achieving the best average accuracy with reasonable token efficiency.These results suggest that different perturbation behaviors lead to different efficiency--accuracy trade-offs.

Overall, all perturbation styles achieve stable and competitive results, indicating that SABER does not rely on a specific manually designed perturbation prompt. Instead, its effectiveness primarily depends on the perturbation behavior induced by the prompt rather than its exact wording.

\subsection{Trigger Sensitivity Analysis}
\begin{table*}[t]
\centering
\small
\setlength{\tabcolsep}{3.5pt}
\begin{tabular}{l cc cc cc cc cc cc | cc}
\toprule
\multirow{2}{*}{\textbf{Trigger}}
& \multicolumn{2}{c}{\textbf{GSM8K}}
& \multicolumn{2}{c}{\textbf{MATH-500}}
& \multicolumn{2}{c}{\textbf{AIME24}}
& \multicolumn{2}{c}{\textbf{AIME25}}
& \multicolumn{2}{c}{\textbf{Olympiad}}
& \multicolumn{2}{c}{\textbf{GPQA-D}}
& \multicolumn{2}{c}{\textbf{Overall}} \\
& Acc$\uparrow$ & Tok$\downarrow$
& Acc$\uparrow$ & Tok$\downarrow$
& Acc$\uparrow$ & Tok$\downarrow$
& Acc$\uparrow$ & Tok$\downarrow$
& Acc$\uparrow$ & Tok$\downarrow$
& Acc$\uparrow$ & Tok$\downarrow$
& Acc$\uparrow$ & CR$\downarrow$ \\
\midrule

Vanilla
& \textbf{96.4} & 2,192
& \textbf{92.8} & 5,135
& \textbf{66.7 }& 12,003
& 54.2 & 12,999
& 59.0 & 9,447
& 55.1 & 9,047
& 70.7 & 100\% \\

\texttt{``Wait''}
& 96.0 & \textbf{1,258}
& 92.4 & \textbf{2,303}
& 65.0 & 9,645
& 52.5 & \textbf{10,592}
& \textbf{61.5} & \textbf{4,892}
& \textbf{57.6} & \textbf{5,056}
& \textbf{70.8} & \textbf{66.4\%} \\

Newline ($\texttt{\textbackslash n\textbackslash n}$)
& 95.7 & 1,312
& \textbf{92.8} & 3,387
& 65.0 & \textbf{9,115}
& \textbf{55.0} & 10,797
& 59.7 & 6,190
& 56.1 & 5,726
& 70.7 & 71.9\% \\

\bottomrule
\end{tabular}
\caption{Trigger sensitivity analysis on Qwen3-8B. Replacing the lexical trigger ``Wait'' with a semantically neutral double newline ($\texttt{\textbackslash n\textbackslash n}$) yields comparable performance, suggesting limited sensitivity to the choice of trigger.}
\label{tab:trigger_sensitivity}
\end{table*}
To assess the sensitivity of SABER to the specific lexical trigger ``Wait'', we replace it with a more general and semantically neutral double newline ($\texttt{\textbackslash n\textbackslash n}$), while keeping all other settings unchanged. As shown in Table~\ref{tab:trigger_sensitivity}, the newline-based trigger achieves nearly identical overall accuracy (70.7\% vs.~70.8\%) and comparable token compression (71.9\% vs.~66.4\%). This suggests that SABER is not strongly tied to the specific lexical cue ``Wait''. The trigger primarily provides a practical way to locate potential reasoning transition points for probe generation.

\section{Case Study}
\label{appendix:case}

We present several representative cases generated by Qwen3-8B under the default SABER setting ($\alpha=0.7, \tau=0.95$). These cases illustrate how SABER adaptively identifies stable reasoning states through adversarial branch probing, thereby reducing unnecessary reasoning and mitigating overthinking.

\begin{figure*}[t]
    \centering
    \includegraphics[width=\linewidth]{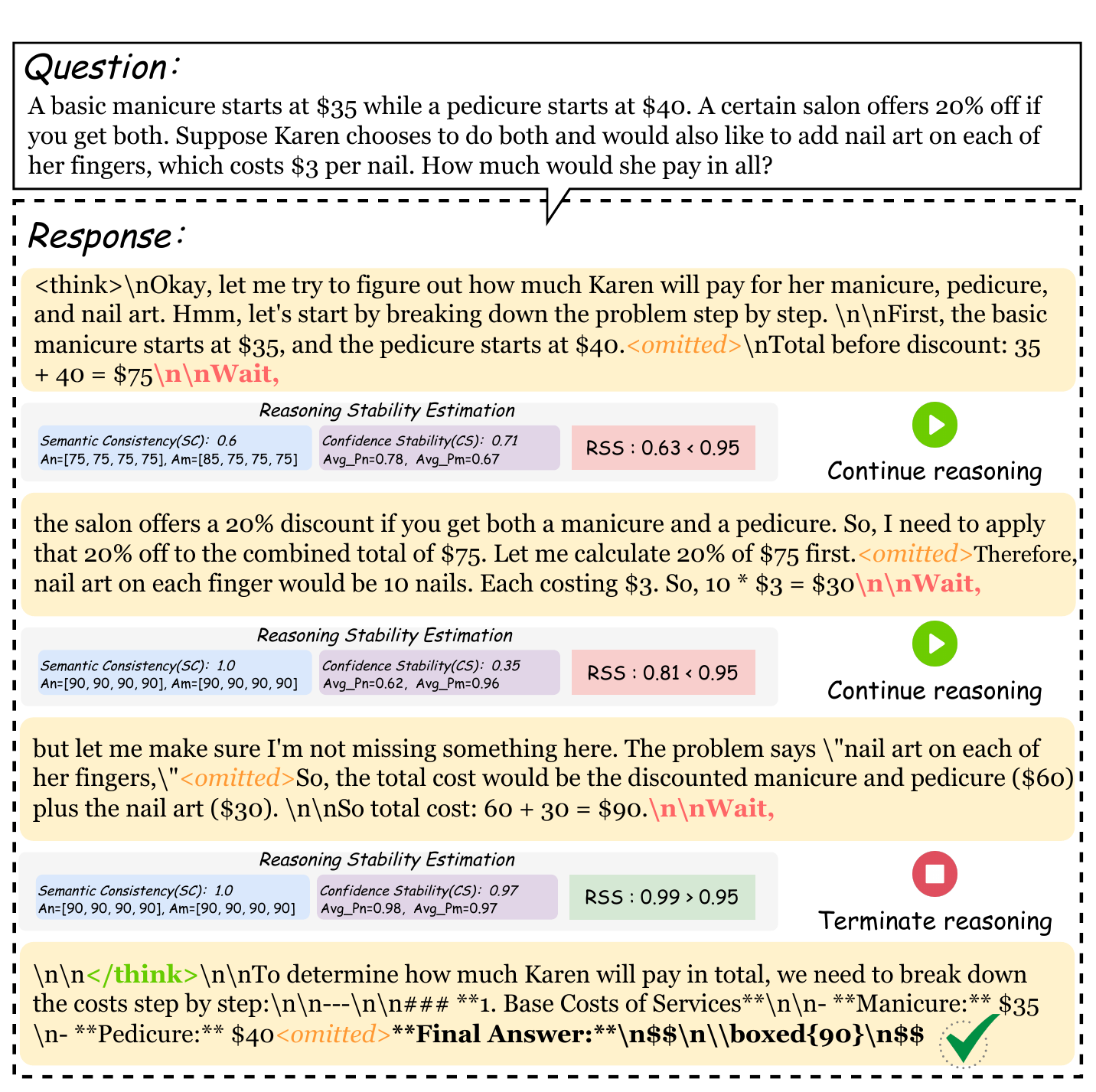}
    \caption{A case on GSM8K where SABER identifies unstable reasoning states.}
    \label{fig:case1}
\end{figure*}

\paragraph{Case 1: Detecting Unstable Intermediate Reasoning States.}
Figure~\ref{fig:case1} illustrates how SABER dynamically evaluates reasoning stability during multi-step mathematical reasoning on GSM8K. In the early reasoning stages, although the neutral branch already produces highly consistent answers, the adversarial branch remains unstable under perturbation, indicating that the intermediate reasoning state has not yet stabilized. Consequently, SABER continues the reasoning process instead of exiting early. As reasoning progresses, both branches gradually stabilize and converge to the same correct answer, causing the RSS to exceed the predefined threshold and trigger early termination. This example shows that local answer agreement or high confidence alone may provide misleading stopping signals, whereas SABER evaluates whether reasoning remains stable under adversarial perturbations to better identify stable reasoning states.

\begin{figure*}[t]
    \centering
    \includegraphics[width=\linewidth]{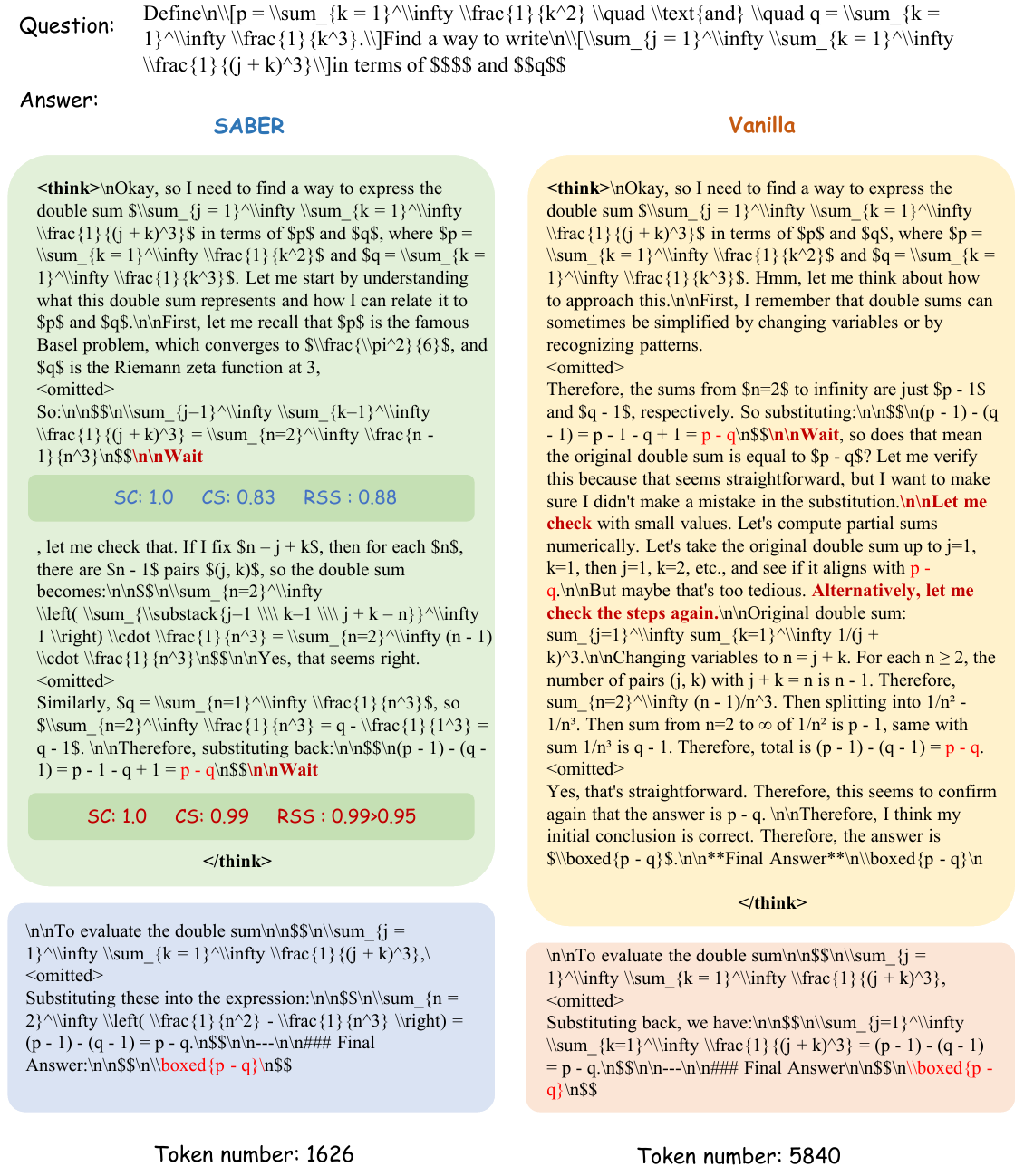}
    \caption{Comparison of vanilla reasoning and SABER on a MATH-500 problem. SABER reduces token consumption from 5840 to 1626 while maintaining correctness.}
    \label{fig:case2}
\end{figure*}

\paragraph{Case 2: Avoiding Redundant Overthinking.}
Figure~\ref{fig:case2} compares SABER with vanilla reasoning. Although both methods ultimately arrive at the correct answer, Vanilla continues to perform repeated verification and reflection after reaching the correct solution, generating substantial redundant reasoning. SABER terminates reasoning once stable convergence is detected, substantially reducing reasoning token consumption.

\begin{figure*}[t]
    \centering
    \includegraphics[width=\linewidth]{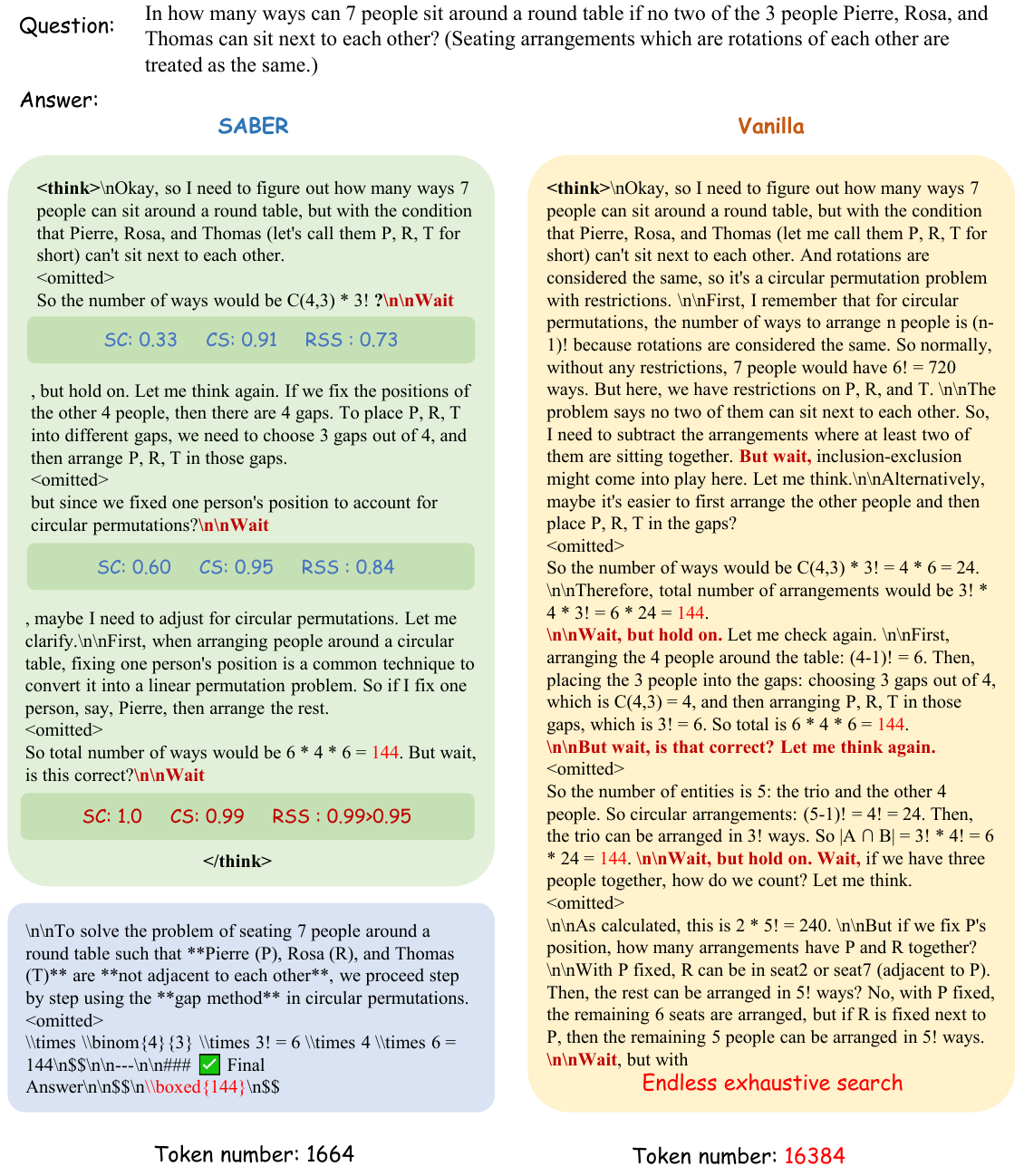}
    \caption{A case on MATH-500 where vanilla reasoning exceeds the context limit due to endless reflection, while SABER exits early and produces the correct answer.}
    \label{fig:case3}
\end{figure*}

\paragraph{Case 3: Preventing Endless Reflection and Reasoning Truncation.}
Figure~\ref{fig:case3} shows a case where vanilla reasoning exceeds the context limit due to persistent reflection. Although the model actually arrives at the correct answer early, it repeatedly verifies and reflects, ultimately failing to output the final result. SABER instead stops reasoning once the reasoning state becomes sufficiently stable, thereby avoiding truncation caused by excessive reflection.

\begin{figure*}[t]
    \centering
    \includegraphics[width=\linewidth]{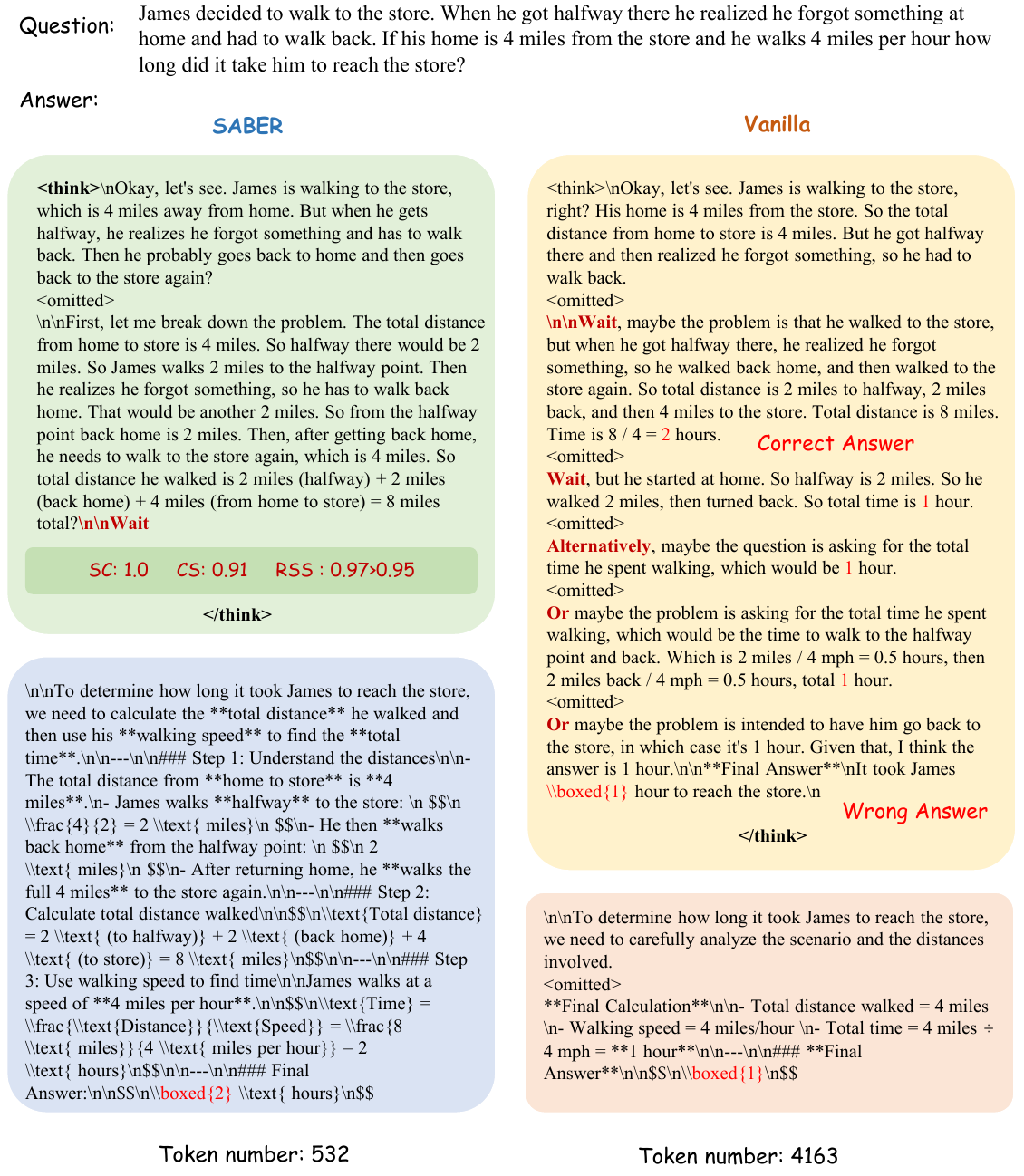}
    \caption{A GSM8K case where vanilla reasoning degrades the correct answer through over-reflection, while SABER preserves it by timely early exit.}
    \label{fig:case4}
\end{figure*}

\paragraph{Case 4: Overthinking Can Hurt Reasoning Accuracy.}
In the case shown in Figure~\ref{fig:case4}, vanilla reasoning arrives at the correct answer early, but during subsequent reflection, it modifies its original conclusion and ultimately produces an incorrect answer. In contrast, SABER avoids unnecessary post-hoc reflection and successfully preserves the correct answer. This case further suggests that excessive reflection may degrade reasoning performance.

\end{document}